\documentclass{article}

\usepackage{arxiv}

\usepackage[utf8]{inputenc} 
\usepackage[T1]{fontenc}    
\usepackage{hyperref}       
\usepackage{url}            
\usepackage{booktabs}       
\usepackage{amsfonts}       
\usepackage{nicefrac}       
\usepackage{microtype}      
\usepackage{lipsum}		
\usepackage{graphicx}
\usepackage{doi}
\usepackage[ruled,lined]{algorithm2e}
\usepackage{bookmark}
\usepackage{appendix}
\usepackage{amsmath}
\usepackage{listings}
\usepackage{xcolor}
\usepackage{amssymb}
\usepackage[numbers]{natbib}

\title{Is Implicit Knowledge Enough for LLMs? A RAG Approach for Tree-based Structures}

\author{{Mihir Gupte}\thanks{Work done during rotation in Connected Vehicle Experience Research, General Motors} \\
	General Motors \\
	\texttt{mihir.gupte@gm.com} \\
	\And
	Paolo Giusto \\
    General Motors \\
	\texttt{paolo.giusto@gm.com} \\
    \And
    Ramesh S. \\
    General Motors \\
    \texttt{ramesh.s@gm.com} \\
}

\renewcommand{\shorttitle}{Technical Report}

\hypersetup{
pdftitle={Is Implicit Knowledge Enough for LLMs? A RAG Approach for Tree-based Structures},
pdfsubject={},
pdfauthor={Mihir Gupte},
pdfkeywords={Retrieval-Augmented Generation},
}

\begin{document}
\maketitle

\begin{abstract}
Large Language Models (LLMs) are adept at generating responses based on information within their context. While this ability is useful for interacting with structured data like code files, another popular method, Retrieval-Augmented Generation (RAG), retrieves relevant documents to augment the model's in-context learning. However, it is not well-explored how to best represent this retrieved knowledge for generating responses on structured data, particularly hierarchical structures like trees. In this work, we propose a novel bottom-up method to linearize knowledge from tree-like structures (like a GitHub repository) by generating implicit, aggregated summaries at each hierarchical level. This approach enables the knowledge to be stored in a knowledge base and used directly with RAG. We then compare our method to using RAG on raw, unstructured code, evaluating the accuracy and quality of the generated responses. Our results show that while response quality is comparable across both methods, our approach generates \textit{over 68\% fewer documents in the retriever}, a significant gain in efficiency. This finding suggests that leveraging implicit, linearized knowledge may be a highly effective and scalable strategy for handling complex, hierarchical data structures.
\end{abstract}


\section{Introduction}
\label{sec:intro}

Large Language Models (LLMs) have become the cornerstone of many applications, from chatbots to advanced analytics. This widespread adoption is fueled by their remarkable ability to learn from in-context information, a property often referred to as "in-context learning" (\cite{dong2022survey}). Furthermore, their proficiency in learning from just a few examples, a concept known as few-shot learning, makes them highly effective for tasks like information retrieval and analysis (\cite{brown2020language}).

A popular and robust method for enhancing LLMs is Retrieval-Augmented Generation (RAG) (\cite{gao2023retrieval}). This technique works by first retrieving a set of relevant documents from a knowledge base in response to a user query. The LLM is then prompted to generate an answer based on both the query and the retrieved knowledge. A significant advantage of RAG is its structure-agnostic nature, allowing it to operate effectively on raw, unstructured data stored as vector embeddings, such as plain text.

While RAG has proven effective, its application to diverse data structures is an active area of research. For instance, studies have explored its use with graph-based data, which is common in social networks and other complex systems. A breadth of recent work (\cite{wang2023can,fatemi2023talk,li2024can}), propose methods for performing RAG over knowledge graphs, explicitly capturing details like adjacency and centrality. While this approach is well-suited for applications where information is naturally represented as a graph, it can be computationally intensive and may be considered overkill for other hierarchical structures, which are often a subset of graphs (e.g., tree-like folder structures).

This brings to light a critical question in RAG: what is the most efficient way to capture knowledge? While LLMs can generate high-quality responses from raw, explicit knowledge, it remains to be seen whether the same quality can be achieved by implicitly capturing that knowledge.

A critical technical step in applying RAG to such structures is linearization, which converts the multi-dimensional, nested hierarchy (like a file tree) into a one-dimensional sequence of discrete documents. These documents, which serve as the chunks of information stored in the vector database, are constrained by a set limit of tokens. Thus, when we propose linearizing the tree by generating implicit knowledge, we are employing a method to create these optimized, token-limited documents for the tree's components by distilling their core content and context, rather than using raw data. An example of generated implicit knowledge can be found in Appendix \ref{append:implicit}.

\begin{figure}[h] 
    \centering
    \includegraphics[width=0.75\linewidth]{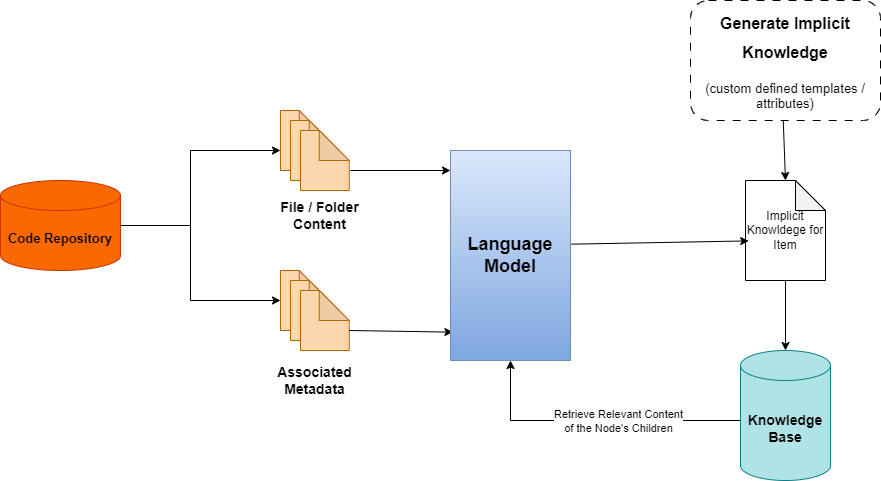} 
    \caption{Workflow for generating Implicit Knowledge from a given set of hierarchical files}
    \label{fig:generate-implicit}
\end{figure}

This paper addresses these intersecting challenges within the context of a common problem: enabling an application to "talk" to unstructured code repositories. Not only are code repositories naturally organized in a hierarchical, tree-like format, but they are also often raw, unstructured, and not explicitly documented.

Our work makes the following contributions:

\begin{enumerate}
    \item We propose a novel method for traversing hierarchical structures like trees to generate implicit knowledge tailored for RAG-type applications.
    \item We demonstrate that our implicit knowledge generation method leads to significantly more efficient RAG usage. Our findings show that our approach yields a comparable quality of responses while generating almost four times less data in the vector database. This suggests that implicit knowledge may be sufficient and more efficient for managing complex, structured information.
\end{enumerate}

\section{Related Work}
\label{sec:relatedwork}

Existing graph-based LLM applications can be broadly categorized into two types. The first category focuses on solving fundamental graph problems, such as those related to node properties (e.g., degree, centrality) and relational properties (e.g., shortest path between two nodes) \cite{wang2023can,fatemi2023talk,li2024can}. The second category, which is more relevant to our work, evaluates the ability of LLMs to "talk to a graph" by answering natural language queries about its content and structure  (\cite{he2024g, sarthi2024raptor}).

Our work shares a conceptual similarity with the latter category, particularly the bottom-up knowledge aggregation approach proposed by  \cite{sarthi2024raptor}. However, our primary objective is to distill knowledge from a hierarchical structure for a more efficient Retrieval-Augmented Generation (RAG) pipeline. In contrast, their method focuses on retrieving similar embedding vectors and summarizing clusters to provide answers, which represents a different approach to a similar problem.

Another line of work centers on the construction of Knowledge Graphs (KGs) from text and using LLMs for subsequent reasoning over these KGs  (\cite{li2024simple,edge2024local}). These methods primarily address the challenge of structuring unstructured text into a graph format, whereas our work begins with an existing hierarchical structure and focuses on creating an optimal representation of its knowledge for efficient retrieval.

Another line of work relevant to our research is knowledge distillation, which focuses on compressing knowledge from a large model into a smaller one. Our concept of generating implicit knowledge shares commonality with this field. For instance, some methods create explicit frameworks, such as using "teacher-models" to improve Chain-of-Thought reasoning  (\cite{deng2023implicit}) or to distill knowledge for unlabeled data  (\cite{wang2023explicit}). Other works use customized loss functions to extract subtler forms of implicit knowledge from a model's activations or internal states (\cite{li2024direct}). While these methods primarily focus on model-level distillation, they underscore the broader principle that valuable knowledge can be distilled into a more compact and efficient form. Our work applies this fundamental idea to the realm of RAG, demonstrating how knowledge can be distilled from a data source itself, rather than from a model. 

Finally, it has been demonstrated that the performance of RAG-based approaches can be significantly affected by the number of documents stored in vector databases, often leading to performance degradation as the context size increases (\cite{levy2025more, warfield2024vector}). This has spurred research into methods that optimize the retrieval process and manage large contexts more effectively, such as the work on long-context RAG by Jiang et al. (\cite{jiang2024longrag}). This body of literature underscores the critical need for developing efficient and scalable methods to manage the knowledge used in RAG frameworks. Our work on generating implicit knowledge addresses this need directly by enabling more concise and effective storage of information, thus mitigating the very performance issues identified in these studies.

\section{Generating Implicit Knowledge for Trees}
\label{sec:methods}

A key challenge with using Retrieval-Augmented Generation (RAG) on hierarchical structures is that simply linearizing a tree and capturing raw data does not account for the holistic semantic information contained within the entire structure. A practical example is a "folder-level" query, where naively storing raw information can lead to a loss of contextual semantics. Additionally, raw data, especially code, can be extremely long-winded and token-heavy, making it inefficient to capture in its entirety.

To mitigate these problems, we propose a novel method to iterate through hierarchical structures to generate \textit{implicit knowledge} on trees. We first traverse to the leaf-level of a given tree and capture the "templated" knowledge using a large language model (LLM). We generate implicit knowledge for all leaf nodes in this manner. Subsequently, we traverse to the parent node of each leaf and generate a higher-level summary by using all the implicit knowledge previously generated for its children. This ensures that the parent node has a holistic understanding of its entire sub-tree. The algorithm for traversing and generating implicit knowledge is given by Algorithm \ref{algo:traverse}.

\begin{algorithm}[H]
    \label{algo:traverse}
    \caption{Bottom-Up Knowledge Aggregation}
    \SetKwInOut{Input}{Input}
    \SetKwInOut{Output}{Output}
    \Input{}
    Tree structure \textbf{$T$} representing the file hierarchy \\
    Large language model \textbf{$L$} \\
    Prompt template for leaf nodes \textbf{$P_{\text{leaf}}$} \\
    Prompt template for parent nodes \textbf{$P_{\text{parent}}$} \\

    \Output{Knowledge base $KB$}
    
    \Begin{
        \ForEach{leaf node $i \in T$}{
            Generate implicit knowledge $ImpKnowledge_i$ using $P_{\text{leaf}}$ and $L$\;
            Store $ImpKnowledge_i$ in $KB$\;
        }
        
        \While{not reached root of $T$}{
            \ForEach{parent node $N$ at current level}{
                Collect $\{ImpKnowledge_i\}$ from all child nodes $i$ of $N$\;
                Generate implicit knowledge $ImpKnowledge_N$ using $P_{\text{parent}}$ and $L$\;
                Store $ImpKnowledge_N$ in $KB$\;
            }
            Move one level up in $T$\;
        }
    }
\end{algorithm}

\begin{figure}[h] 
    \centering
    \includegraphics[width=1\linewidth]{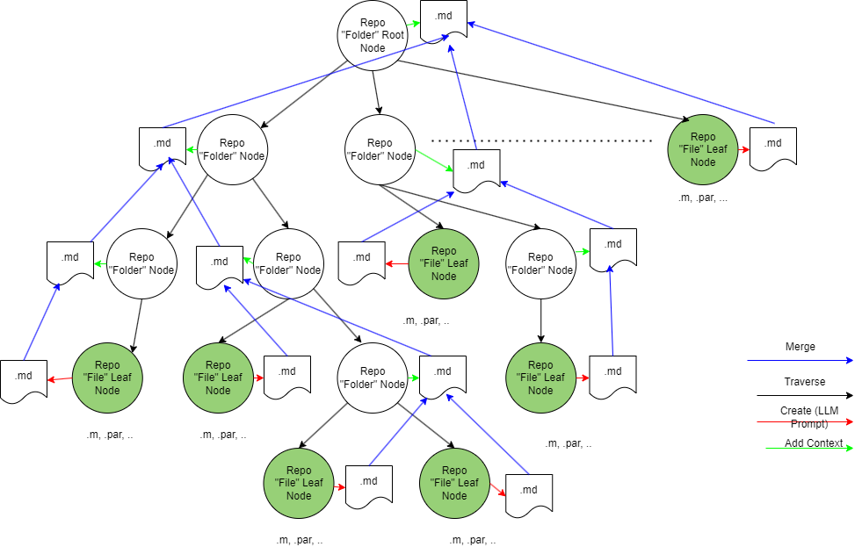} 
    \caption{Workflow for generating Implicit Knowledge from a given set of hierarchical files}
    \label{fig:sample-traversal}
\end{figure}

The specific content of this template can vary depending on the use case for a given hierarchical structure; the template we utilized is provided in Appendix \ref{append:prompt}.

\section{Experiment Setup}
\label{sec:data_experiment}

\subsection{Data Setup}
For our experiments, we use a realistic automotive scenario of an unstructured code repository. The repository holds various files, predominantly MATLAB Simulink scripts designed for vehicle simulation in different types of environments. The data is particularly challenging because it does not represent a single, organized software package. Instead, it comprises a diverse set of utilities and simulation files collected by a team over several years, often stored in a decentralized and ad-hoc manner. While the files themselves are arranged in a hierarchical (tree-like) structure, the repository as a whole lacks a cohesive, logical organization, making it an ideal real-world test case for our method.

\subsection{Setup of Methods}
We compare two distinct methods. Our \textit{baseline method} establishes a conventional RAG pipeline by directly indexing and storing the raw content of the code files and their associated GitHub metadata (e.g., file paths, names) in a vector database.

For our \textit{proposed method}, we first generate implicit knowledge by programmatically traversing the repository using the novel approach detailed in Section \ref{sec:methods}. This generated knowledge, rather than the raw files, is then stored and indexed in the vector database.

\subsection{Qualitative Evaluation}

We qualitatively constructed a small yet representative test dataset consisting of 10 questions with corresponding ground-truth answers. To establish the fidelity of the evaluation standard, the ground-truth answers were rigorously defined and validated by Subject Matter Experts (SMEs). The SMEs ensured that each answer was factually correct, complete, and accurately represented the information contained within the code repository. The questions were designed to probe the system's ability to retrieve and synthesize information at both the file-level and folder-level of the hierarchy, and the final results were then quantitatively evaluated using the set of metrics detailed in Section \ref{eval:metric}. A sample question-answer pair is presented below:
\begin{quote}
    \textit{Question: Is there a file that performs Aerodynamic simulation over Track and Coast data?}
    
    \textit{Answer: Yes, the file \texttt{AeroMapCompare.m} plots "Track" and "Coast" data, representing different driving scenarios. It specifically visualizes various ride height metrics in millimeters using subplots.}
\end{quote}
For a complete list of questions, please refer to Appendix \ref{append:list}.

\subsection{Evaluation Metrics}
\label{eval:metric}
We evaluate both methods using three popular string-matching metrics to measure how closely the LLM-generated answers align with the ground-truth answers. These metrics are commonly used in natural language generation and summarization tasks (\cite{gao2023retrieval}).

\subsubsection{BLEU-1}
The \textit{BLEU-1} (Bilingual Evaluation Understudy) score is a metric used to evaluate the quality of text generated by an LLM by comparing it to one or more human-written reference texts. The score ranges from 0 to 1, with a higher score indicating a greater degree of similarity between the generated and reference texts.

The BLEU-1 score is a precision-based metric that measures the overlap of unigrams between a candidate (generated) text and a set of reference texts. Its formula is a modified form of unigram precision.

The core of the BLEU-1 score is calculated as follows:
$$ \text{BLEU-1} = \text{BP} \cdot \exp \left( \sum_{n=1}^{1} w_n \log p_n \right) $$
Where:
\begin{itemize}
    \item $BP$ is the brevity penalty, a factor that penalizes short generated sentences. It's calculated as:
    $$ BP =
    \begin{cases}
        1 & \text{if } c > r \\
        e^{(1 - r/c)} & \text{if } c \le r
    \end{cases} $$
    where $c$ is the length of the candidate sentence and $r$ is the effective reference corpus length.
    \item $w_n$ is the weight for each n-gram. For BLEU-1, we only consider unigrams, so $w_1 = 1$ and all other weights are 0.
    \item $p_n$ is the modified n-gram precision. For BLEU-1, this is $p_1$, the modified unigram precision.
\end{itemize}

\subsubsection{E-F1 Score}

The \textit{E-F1} (Extraction F1) score is a metric that evaluates the token-level overlap between a model's generated answer and a reference answer. It's a more granular measure than BLEU-1, focusing on how well the model extracts the key entities or phrases from the source text.

\begin{itemize}
    \item \textbf{Precision ($P$):} This metric answers the question, "Of all the tokens the model generated, how many were correct?" It is calculated as the number of correct tokens divided by the total number of tokens in the generated answer.
    \begin{equation*}
        P = \frac{\text{Number of Correct Tokens}}{\text{Total Tokens in Generated Answer}}
    \end{equation*}

    \item \textbf{Recall ($R$):} This metric answers the question, "Of all the tokens in the reference answer, how many did the model correctly generate?" It's the number of correct tokens divided by the total number of tokens in the reference answer.
    \begin{equation*}
        R = \frac{\text{Number of Correct Tokens}}{\text{Total Tokens in Reference Answer}}
    \end{equation*}
\end{itemize}

The E-F1 score combines precision and recall into a single metric, providing a balanced measure of both correctness and completeness. The formula is:
\begin{equation*}
    F_1 = 2 \cdot \frac{P \cdot R}{P + R}
\end{equation*}

The score works by counting the number of common tokens between the generated and reference answers. It then calculates precision and recall based on this count, and finally, it computes the harmonic mean. A higher F1 score indicates a better balance between not generating irrelevant words (high precision) and generating all the necessary words (high recall) from the reference.

\subsubsection{EM Formula}
The formula for \textit{token-level EM} is based on a simple ratio of matching tokens. It is calculated by dividing the number of tokens in the reference that are found in the prediction by the total number of tokens in the reference.
\begin{equation*}
    \text{EM}_{\text{token}} = \frac{\text{Number of reference tokens found in prediction}}{\text{Total number of tokens in reference}}
\end{equation*}

\section{Results}
\label{sec:results}
The results of our comparative analysis are presented in Table \ref{tab:results}. The table shows the mean scores for three evaluation metrics---BLEU-1, E-F1, and EM---across file-level, folder-level, and overall evaluation tasks for both the baseline and our proposed method.

Table \ref{tab:results} summarizes the results for the various metrics described in the previous section. We also calculate the amount of documents generated in the vector database in Table \ref{tab:documents}.
\paragraph{}
\begin{table}[h]
    \centering
    \begin{tabular}{|l|ccc|ccc|ccc|}
        \hline
        & \multicolumn{3}{c|}{\textbf{File}} & \multicolumn{3}{c|}{\textbf{Folder}} & \multicolumn{3}{c|}{\textbf{Overall}} \\
        \cline{2-10}
        & \textbf{Bleu-1} & \textbf{E-F1} & \textbf{EM} & \textbf{Bleu-1} & \textbf{E-F1} & \textbf{EM} & \textbf{Bleu-1} & \textbf{E-F1} & \textbf{EM} \\
        \hline
        \textbf{Baseline} & \textbf{0.21} & \textbf{0.33} & \textbf{0.61} & 0.13 & 0.23 & 0.54 & 0.17 & 0.28 & 0.58 \\
        \textbf{Proposed} & \textbf{0.21} & \textbf{0.33} & 0.60 & \textbf{0.15} & \textbf{0.28} & \textbf{0.81} & \textbf{0.18} & \textbf{0.30} & \textbf{0.71} \\
        \hline
    \label{tab:results}
    \end{tabular}
    \caption{Results categorized across File, Folder \& Overall questions across our metrics}
\end{table}

As shown in Table \ref{tab:results}, the proposed method consistently performs similar to or outperforms the baseline on the E-F1 and EM metrics across all evaluation levels. This finding indicates that our approach, by generating concise and semantically focused implicit knowledge, is highly effective at extracting the key entities and phrases required for accurate question answering.

In addition to its superior performance on key token-level metrics, our proposed method also demonstrates substantial gains in efficiency. By generating implicit knowledge, our approach requires over 3.5x fewer documents in the vector database compared to the baseline method. This result indicates that implicit knowledge may be a more resource-efficient and scalable solution for managing complex, hierarchical structures.

\begin{table}[h]
    \centering
    \begin{tabular}{|l|c|}
        \hline
        & \textbf{\# of documents generated} \\
        \hline
        Baseline & 490 \\
        \hline
        Proposed & \textbf{156} \\
        \hline
    \label{tab:documents}
    \end{tabular}
    \caption{Number of documents generated for the vector database by each approach}
\end{table}

\begin{figure}[h] 
    \centering
    \includegraphics[width=0.75\linewidth]{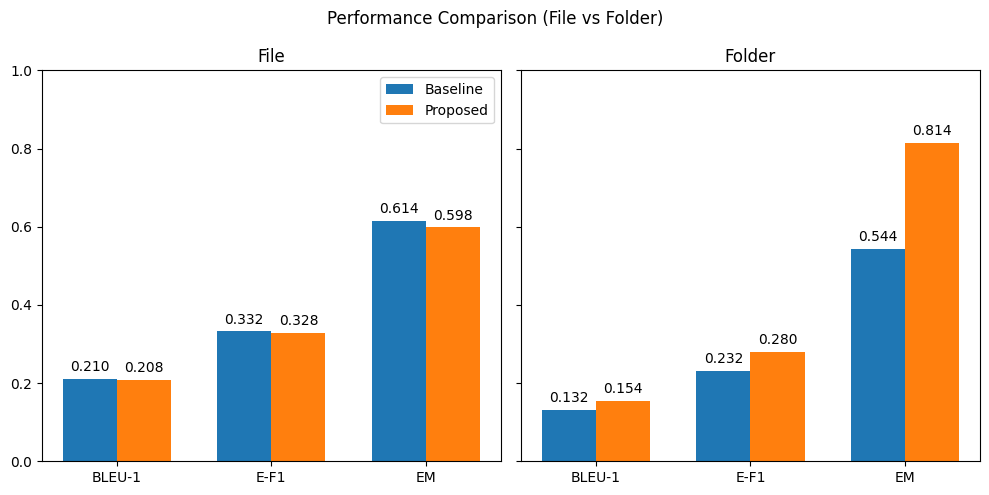} 
    \caption{Performance of baseline \& proposed methods on File-level and Folder-level questions}
    \label{fig:performance}
\end{figure}

\section{Discussion and Limitations}
\label{sec:discussion}

Our experiments provide compelling evidence for the efficacy of using a bottom-up knowledge aggregation approach on hierarchical structures. Our findings demonstrate that generating and utilizing implicit knowledge for RAG tasks not only yields comparable or superior performance but also offers significant gains in efficiency.

\subsection{Implicit Knowledge vs. Raw Data}

As our experiments on code repositories show, models operating on implicit knowledge, which is distilled from raw data, achieve a higher quality of response on key token-level metrics. Specifically, our method demonstrates superior performance for "folder-level" questions, which require a holistic understanding of information across multiple files. This suggests that the LLM-generated summaries effectively capture the contextual and semantic relationships that are often lost when raw, verbose files are simply linearized.

Furthermore, this process proved to be substantially more efficient. Our method generates almost four times fewer documents in the vector database while maintaining performance parity. This result directly addresses a major limitation of RAG pipelines, where an increasing number of documents can lead to performance degradation. By providing the model with distilled, relevant context, our approach optimizes RAG usage and leads to a more scalable solution.

However, these findings also raise several important open questions for future research:
\begin{enumerate}
    \item Can this implicit knowledge generation method be generalized to other non-hierarchical structures, such as knowledge graphs or relational databases?
    \item Is there an optimal representation for implicit knowledge that balances efficiency and performance for a given LLM? In this paper, we only explored a simple Markdown-based structure.
    \item Is linearization, even with our novel aggregation method, the most effective way to capture information from hierarchical structures?
\end{enumerate}

\subsection{Generalizing to Graph-Based Tasks}

A significant body of literature evaluates Graph-based RAG tasks on structural questions pertaining to relationships, such as degree and centrality (\cite{wang2023can,fatemi2023talk,li2024can}). As a future direction, we propose exploring whether our implicit knowledge generation approach can be adapted to these more complex data structures (\cite{ren2024survey}). While a direct hierarchical traversal may not be applicable to arbitrary graphs, the underlying idea of creating holistic, aggregated knowledge from interconnected nodes could still be effective. We would ideally explore benchmark datasets used in graph-based RAG to evaluate if generating implicit knowledge improves performance on these tasks.

\subsection{Implicit Knowledge as a General Paradigm}

Our work provides a proof-of-concept that LLMs do not require raw data; they simply require high-quality implicit knowledge. We have demonstrated that this leads to equal, if not superior, performance while utilizing a significantly smaller amount of data. This finding helps address the challenge of performance degradation that LLMs often face when provided with larger amounts of context (\cite{levy2025more, shi2023large}). Our experiments, however, are limited to a specific code repository scenario.

A key question for future work is to investigate the generalizability of this paradigm: will generating implicit knowledge lead to similar performance for a broader range of tasks and domains? We also aim to formally quantify the trade-off between the level of implicit knowledge and the final performance of the RAG pipeline.

\section{Conclusion}
\label{sec:conclusion}

In this work, we introduced a novel approach to optimizing Retrieval-Augmented Generation (RAG) pipelines for hierarchical data, with a focus on unstructured code repositories that naturally form tree-based structures. Instead of flattening the hierarchy, our method applies a bottom-up traversal and aggregates implicit knowledge from the tree, providing a compelling alternative to traditional RAG.
Our results highlight two major advantages over the baseline: \textit{reduced semantic loss} and \textit{greater efficiency}. First, our implicit knowledge approach produced responses of comparable or superior quality on key token-level metrics, showing that high-level knowledge can be sufficient for both file-level and folder-level queries. Second, our method was significantly more efficient, requiring over 3.5x fewer documents in the vector database.
This study opens several promising avenues for future research, including extending the implicit knowledge paradigm to other structured domains such as graphs, and further exploring the balance between knowledge distillation and model performance. Ultimately, our findings suggest that structure-aware RAG methods can provide a path toward more accurate, efficient, and scalable retrieval systems.

\lstset{
  basicstyle=\ttfamily\small,
  breaklines=true,
  frame=single,
  backgroundcolor=\color{gray!10},
  keywordstyle=\color{blue},
  numbers=none,
  xleftmargin=0pt,
  framexleftmargin=0pt,
  literate={\\land}{{$\land$}}1
           {\\lnot}{{$\lnot$}}1
           {\\neq}{{$\neq$}}1
}
\newpage
\begin{appendices}
    \section{Prompt Template}
    \label{append:prompt}
    \subsection{Template for Generating Root-level (file-level) Knowledge}
    \begin{lstlisting}[language=]
## Task Overview
You task is to generate a .md file from a single file (e.g., .m, .par, .py, etc.) that includes a script that
may include functions defined with related I/O variables and invoked in the script, parameters, 
and classes defined with methods used to create objects of the class, and perform other operation of objects of the class.

Generate metadata for the asset using the following information that the asset contains:
```{context}```

Use the below information to generate the metadata:
If the  file appear to be a script or a function (e.g., .m, .py, etc..), consider the following classification of a script's maturity level when providing your answer:
    1. Ad-hoc/Initial Stage:
        a. Description: 
            These are scripts created for immediate, specific tasks. They often lack formal structure, documentation, and error handling.  
            They might be "one-off" solutions.
            Very similar to a prototype stage.
        b. Characteristics: Minimal or no comments.
            Limited error checking.
            Often not version-controlled.
            Highly dependent on the creator's knowledge.
    2. Proof of Concept/Experimental Stage:
        a. Description:
            Scripts developed to validate a specific idea or approach.
            They demonstrate functionality but may not be optimized for performance or reliability.
            This is where the script is being tested for viability.  
        b. Characteristics:
            Basic documentation.
            Some level of error handling.
            May be subject to frequent changes.
            Used for testing and experimentation.  
    3.  Development/Refinement Stage:
        a. Description:
            Scripts that are actively being improved and refined.
            They start to incorporate better coding practices, error handling, and documentation.  
            This is where the scripts start to become more stable.
        b. Characteristics:
            Improved documentation and comments.
            Robust error handling and logging.
            Version control implementation.
            Increased code modularity and reusability.
    4. Stable/Production Stage:
        a. Description:
            Scripts that are considered reliable and ready for regular use.
            They have undergone thorough testing and are well-documented.
            They are integrated into established workflows.
        b. Characteristics:
            Comprehensive documentation.
            Extensive error handling and logging.
            Rigorous testing and validation.
            Integration with other systems and processes.
    5. Optimized/Mature Stage:
        a. Description:
            Scripts that are continuously monitored and optimized for performance, efficiency, and maintainability.  
            They are subject to ongoing improvements and updates.
            Similar to the CMMI optimizing level.
        b. Characteristics:
            Performance monitoring and optimization.
            Automated testing and deployment.
            Continuous integration and continuous delivery (CI/CD).
            Proactive maintenance and updates.
It's important to note that the maturity of a script or a function invoked in a script is often context-dependent. 
So, if you can try to infer the context of the script. For a simple script used for personal tasks may never need to reach the "optimized" stage, 
while a script used in a critical production environment will require a high level of maturity.
    \end{lstlisting}
    \subsection{Template for Generating Parent-level (folder-level) Knowledge}
    \begin{lstlisting}[language=]
## Task Overview
You task is to generate a .md file for the folder, by merging several .md files, each representing an item in the folder.
It is possible that some .md files included in the folder are describing .m, .par, .py files while other .md files are describing folders included
in this folder. 
When you generate the single .md file please consider if the contents describe in the .md files (one per item in the folder) are somehow
interrelated or just a loose collection of things with a similar them.

Generate metadata using the following information:
```{context}```

## Input 
A set of files with .md extension each file summarizing an item in the folder
## Output Requirements
Generate a structured meta-description with the following sections:
1. Folder Name
2. Folder Location in the repo (tree-structure)
3. An aggregated summary of the folder contents
Here is an example of the summary: 
"This folder contains data and scripts related to the analysis of a Porsche Taycan 
Turbo S's lap record at the Nurburgring. The primary focus is on extracting and analyzing 
vehicle speed and lap time data from video footage of the lap. 
The analysis involves several steps, including video pre-processing, 
frame extraction, optical character recognition (OCR) for data extraction, 
and data plotting for insights into vehicle performance."
4. An list of the folder types of contents (e.g., data, scripts, functions, classes, models, etc..)
5. A more detailed description of the folder contents. If the folder has scripts, describe the scripts one by one,
if the folder has data describe the data, etc...
6. A set of references (e.g., from tools documents that are used to run the script)
7. The folder creator name
8. The folder creation date
    \end{lstlisting}
    \newpage
    \section{List of Questions}
    \label{append:list}
    
    \textbf{File-level Questions}
    \begin{enumerate}
        \item \textit{What does the file Lap\_Video\_to\_Speed do?}
        \item \textit{Is there a file that performs Aerodynamic simulation over Track and Coast data?}
        \item \textit{What is the name of the file that measures Brake Rotor Temperature as a function of time?}
        \item \textit{What parameters are used to plot the Motor Torque in MotorCurvePlotTool.m?}
        \item \textit{What symbols are used to calculate the peak tire slip based on MF-Tire?}
    \end{enumerate}
    \textbf{Folder-level Questions}
    \begin{enumerate}
        \item \textit{What information does the folder CarSim CompTire contain?}
        \item \textit{Are there any scripts to evaluate Suspension Tools?}
        \item \textit{Do any scripts make use of the file Read\_TIR\_2CompTire\_Func.m?}
        \item \textit{What is the folder LapTimeBenchmarkPlotter folder plotting?}
        \item \textit{What is the difference between the folder for SuspentionDoeCalculator and RollCenterModifier?}
    \end{enumerate}

    \section{Example of Generated Implicit Knowledge}
    \label{append:implicit}
    \subsection{Generated Implicit Knowledge}
    \begin{lstlisting}[language=]
# Metadata for File: FILE_INFORMATION

0. **Name of File:** Core.m

1. **Creator or Author Name and Information**
   - **Author Name:** XXX
   - **Author Email:** XXX

2. **Explanation:** The metadata generated is based on the content and structure of the provided MATLAB class file, `Core.m`. The analysis focuses on the class definition, properties, methods, and the context provided by the latest commit details. The classification of the script's maturity level is inferred from the coding practices, documentation, error handling, and the overall structure observed in the script.

3. **Type of Asset:** This is a class.

4. **A summary of I/Os required to run**
   - **Required Parameters/Objects to Instantiate the Class:**
     - Optional arguments for configuration settings such as 'driverSettingsFile', 'targetPathObjectFile', 'vehicleSettingsFile', 'optSettingsFile', 'resultsInterface', 'simInterface', 'rootModel'.
   - **Example of Instantiation:**
     ```matlab
     SAMPLE MATLAB CODE
     ```

5. **Compatibility:** The class is designed to work within the MATLAB environment, specifically tailored for lap time simulation and optimization tasks. It likely interfaces with other MATLAB scripts, functions, and toolboxes relevant to vehicle dynamics simulation.

6. **Category:** Vehicle Dynamics Simulation and Optimization

7. **Description and Features:**
   - The class provides functionalities for loading configuration settings, resetting iterations, cleaning paths, creating new iterations, creating speed plans, finding the best candidate, loading candidates, running iterations, and optimizing lap times. It also includes private methods for internal operations such as calculating tire force margins, finding failed segment indices, and creating composite best plans.

8. **Keywords:** Lap Time Simulation, Vehicle Dynamics, Optimization, MATLAB, Simulation Interface

9. **Maturity Level:** Development/Refinement Stage
   - **Explanation:** The `Core` class demonstrates characteristics of being in the development/refinement stage. It includes improved documentation and comments throughout the class, robust error handling, and a structured approach to defining properties and methods. The class is designed with modularity in mind, allowing for easy integration and reuse within the framework. The presence of input parsing, default settings, and the ability to handle various configuration options further indicate a move towards a more stable and refined codebase. However, the class may still be subject to changes and improvements as it is actively developed and refined for better performance, reliability, and integration with other systems or processes.
    \end{lstlisting}
\end{appendices}
\newpage
\bibliographystyle{unsrtnat}
\bibliography{template}  






\end{document}